\documentclass{article}

\usepackage[final]{neurips_2022}

\usepackage[utf8]{inputenc} % allow utf-8 input
\usepackage[T1]{fontenc}    % use 8-bit T1 fonts
\usepackage[hidelinks]{hyperref}       % hyperlinks
\usepackage{url}            % simple URL typesetting
\usepackage{booktabs}       % professional-quality tables
\usepackage{amsfonts}       % blackboard math symbols
\usepackage{nicefrac}       % compact symbols for 1/2, etc.
\usepackage{microtype}      % microtypography
\usepackage{xcolor}         % colors
\usepackage{graphicx}

\usepackage{amsmath}
\usepackage{amssymb}
\usepackage{mathtools}
\usepackage{amsthm}
\usepackage{bbm}
\usepackage{caption}
\usepackage{subcaption}

\DeclareMathOperator*{\argmin}{arg\,min}

\title{Improving self-supervised representation learning via sequential adversarial masking}

\author{%
  Dylan Sam \\
  Machine Learning Department\\
  Carnegie Mellon University\\
  \texttt{dylansam@andrew.cmu.edu} \\
   \And
    Min Bai, \; Tristan McKinney, \; Li Erran Li \\
    Amazon Web Services \\
    Amazon \\
    \texttt{\{baimin, tristamc, lilimam\}@amazon.com}
}

\begin{document}

\maketitle

\begin{abstract}

Recent methods in self-supervised learning have demonstrated that masking-based pretext tasks extend beyond NLP, serving as useful pretraining objectives in computer vision. However, existing approaches apply random or ad hoc masking strategies that limit the difficulty of the reconstruction task and, consequently, the strength of the learnt representations. We improve upon current state-of-the-art work in learning adversarial masks by proposing a new framework that generates masks in a sequential fashion with different constraints on the adversary. This leads to improvements in performance on various downstream tasks, such as classification on ImageNet100, STL10, and CIFAR10/100 and segmentation on Pascal VOC. Our results further demonstrate the promising capabilities of masking-based approaches for SSL in computer vision.
\end{abstract}

\section{Introduction}

Self-supervised learning (SSL) involves designing pretext tasks to learn useful representations from unlabeled data. In NLP, prior works \cite{kenton2019bert, Joshi2020SpanBERTIP, Lan2020ALBERTAL} have demonstrated that filling in removed words or sequences of words in a sentence serves as a useful pretext task, learning better representations for various downstream tasks. In computer vision, however, the prevalent trend has been to employ contrastive approaches \cite{chen2020simple}, given that we do not have natural objects to mask out (such as words in sentences). Recent works, such as \cite{pathak2016context, Bao2022BEiTBP, he2022masked, assran2022masked}, have applied masking-based pretext tasks to computer vision tasks and achieved comparable performance to contrastive approaches. One major caveat of these works is that they employ simple, random masking procedures to remove parts of an image. This pretext task is much easier, as the network can perform reconstruction by extending textures or colors. It appears that networks can learn trivial correlations instead of the global structure (over the distribution) of images. This is supported by the boosted performance of MAE \cite{he2022masked} when removing larger amounts of the image, forcing the network to use some notion of global reasoning. 

Our work addresses this masking procedure. We aim to learn \textit{how} to mask out regions of an image to improve upon existing ad hoc masking pipelines. Other recent work looks to learn more meaningful masking procedures \cite{shi2022adversarial, Li2022SemMAESM}. We further build on the initial results of the Adversarial Inference-Occlusion Self-supervision (ADIOS) model \cite{shi2022adversarial}, which learns to generate these masks in an adversarial fashion. This prior work demonstrates that a masking network trained to maximize the error of an encoder can learn to generate imperfect yet semantically meaningful masks. Furthermore, the approach learns better features for downstream tasks when added to the standard contrastive setup. However, there are a few limitations to this existing approach. It simultaneously generates a fixed number of masks $N$, constraining them to be roughly equal in strength through a pixel-wise softmax and a penalty term. We argue that this particular constraint is not flexible and leads to undesirable outcomes. For example, each pixel must be covered by a mask, and these masks are encouraged to be completely disjoint. Moreover, the simultaneous generation by the masking network is somewhat ill-suited to the task as the resulting masks are equally valid when permuted, leading to potentially unstable results. 
% \editcomment{Might want to clarify why these three things are undesirable. Also, I think the problem is that they *are* valid under permutations, while we have the intuition that we want to break that symmetry. -T}
%In addition, when these masks are generated at once, there is no particular encouragement for individual parts to fall into a particular one of the $N$ masks. 

We fundamentally change this masking procedure and provide a new framework that generates masks in a sequential fashion under a different constraint on the adversary. We add a budget parameter $b$ that controls the strength of each of the $N$ masks by requiring each to remove only $b$\%  of the total number of pixels. We note that this is more flexible than the ADIOS framework, which has an implicit budget of $\frac{1}{N}$, as each mask is of roughly equal strength. This allows us to impose the constraint for each mask independently and hence generate the masks sequentially. To encourage the learnt masks to capture different regions of the image, we introduce an overlap penalty (i.e., a dot product of the proposed mask and the sum of all previous masks). This mirrors similar intuition in instance segmentation \cite{Liu_2017_ICCV}, where we remove object proposals one by one. Here, we learn to iteratively mask out a region of the image and pass previously masked regions back into the network, so it selects from the remaining portion of the image. This sequential formulation breaks the symmetry that masks are equally valid under permutations in previous work.

\begin{figure}
    \centering
    \begin{subfigure}[t]{0.415\textwidth}
        \centering
        \includegraphics[width=\textwidth]{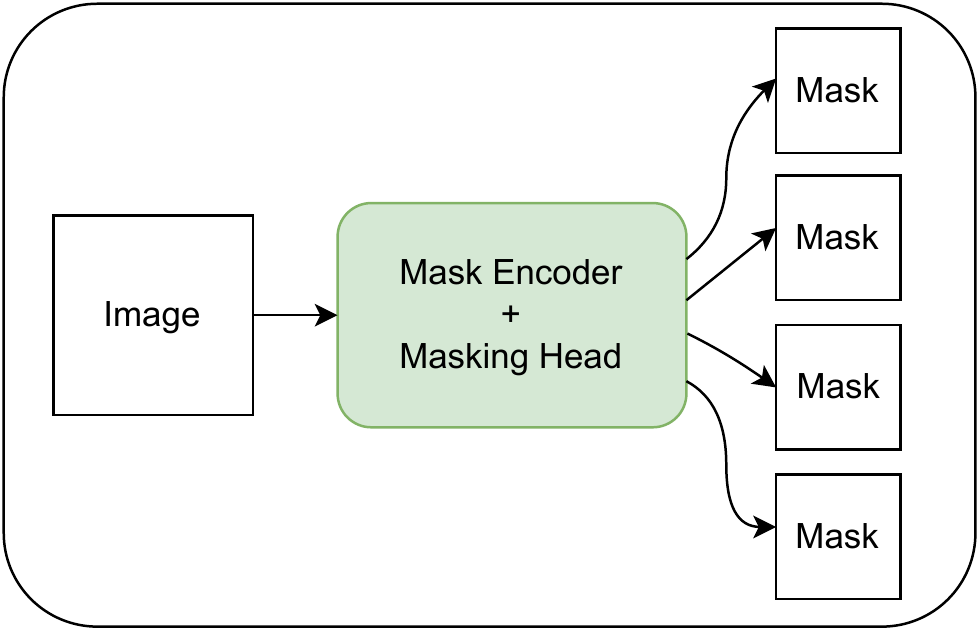}
        \caption{ADIOS framework}
    \end{subfigure}
    \hfill
    \begin{subfigure}[t]{0.55\textwidth}
        \centering
        \includegraphics[width=\textwidth]{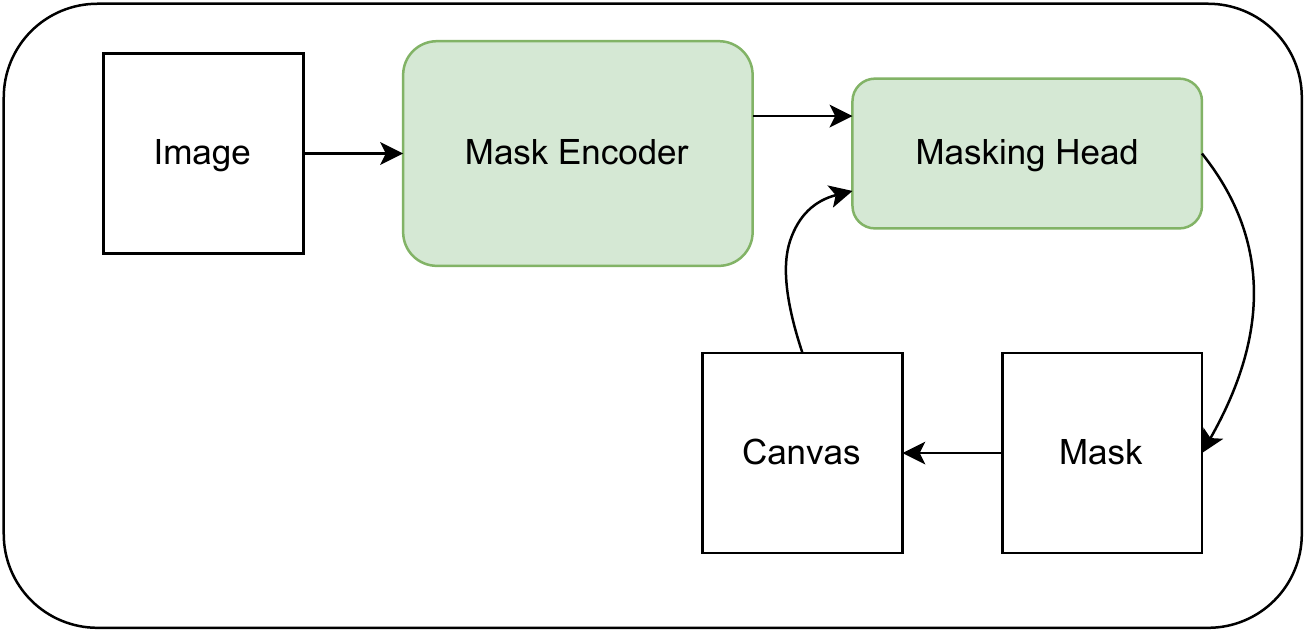}
        \caption{Our sequential masking framework}
        % \editcomment{Can we redraw this diagram a little to show that the output of the masking head is written into the canvas as a loop?}
    \end{subfigure}
    
    \caption{Visualization the masking pipeline $\mathcal{M}$ of existing work (a) and our sequential approach (b). (a) generates masks simultaneously, while (b) feeds masks recurrently into the masking head. }
    \vspace{-3mm}
    \label{fig:masks}
\end{figure}

We apply our pretraining approach to multiple datasets (ImageNet100s, STL10) and demonstrate its efficacy by using the pretrained models for downstream tasks including in distribution (ID) classification, transfer learning, and semantic segmentation on Pascal VOC \cite{everingham2010pascal}. Our method achieves better performance than prior work, especially on linear readout on ImageNet100s (an increase of 3 accuracy points or a 7\% reduction in error).
% \editcomment{can we further clarify this a little, e.g. relative improvement, reduction of error, etc?}. 
We achieve comparable performance in fine-tuning and almost a 5\% relative improvement in mIoU for semantic segmentation on Pascal VOC. We also compare visualizations of our masks  with those generated by existing work in Appendix \ref{app:vis_mask}.

% add a bit more discussion about ongoing or future work !

% \section{Methods}
\section{Preliminaries}

Following the promising results of adversarial approaches in generating mask-based pretext tasks for SSL \cite{shi2022adversarial}, we consider the same setting where we have an encoder model $\mathcal{I}$ and a masking network $\mathcal{M}$, which is illustrated in Figure \ref{fig:masks}. Our masking network generates a set of real-valued masks $m = \mathcal{M}(x)$, which we apply to the original image through a Hadamard product $m \circ x$. We perform our reconstruction in latent space, as is done in ADIOS and other existing work \cite{assran2022masked}. We learn our inference and masking models in an adversarial fashion, captured by the equation 
\begin{equation*}
    \mathcal{I}^*, \mathcal{M}^* = \argmin_{\mathcal{I}} \max_{\mathcal{M}} \mathcal{L}(x, \mathcal{I}, \mathcal{M}),
\end{equation*}
where $\mathcal{L}$ is some loss function. We perform alternating update steps to the encoder model and the masking network, following the procedure in generative adversarial networks \cite{gans}. First, we find the loss of our encoder model $\mathcal{I}$. We use SimCLR \cite{chen2020simple} as our underlying SSL objective, and we denote our positive pairs of $x_i$ as $x_i^A, x_i^B$, where $A, B$ are two randomly sampled augmentations. Then, the loss function of our encoder model is given by
\begin{equation}\label{eq:simclr}
    \mathcal{L}_{\text{SimCLR}}^{\text{encoder}}(x, \mathcal{I}, m) = \log \Big( \frac{\exp(D(\mathcal{I}(x_{i}^A) , \mathcal{I}(x_{i}^B \circ m)}{\sum_{i \neq j} \exp(D(\mathcal{I}(x_{i}^A) , \mathcal{I}(x_{i}^B \circ m))} \Big),
\end{equation}
where $D$ is the negative cosine similarity and $m$ is a mask generated by the adversarial masking network. This is simply the standard SimCLR loss function, except with one of the positive pairs having applied the mask generated by our masking network. 

\section{Sequential Masking Network}

We now propose our new sequential masking framework to improve on prior work that generates masks simultaneously. We describe the loss function of our masking network $\mathcal{M}$, which is depicted in Figure \ref{fig:masks} (b). In Equation \ref{eq:simclr}, we considered the encoder loss for a single mask $m$. However, we generate a set of masks $\{m^1, ..., m^N \}$ for each image $x$. As these are produced in a sequential fashion, we have that $m^k = \mathcal{M}(x, m^1, ..., m^{k-1})$. Given this set of $N$ masks, we compute the \textit{average} loss over all masks, which is also done in ADIOS. This gives us that our masking network loss is given by
\begin{equation*}
    \mathcal{L}_{\text{SimCLR}}^{\text{mask}} = \frac{1}{N} \sum_{i=1}^N \mathcal{L}_{\text{SimCLR}}^{\text{encoder}}(x, \mathcal{I}, \mathcal{M}(x, m^1, ..., m^{i-1})).
\end{equation*}
If $\mathcal{M}$ is unconstrained, then it would trivially learn to remove all the pixels from an image. Therefore, we incorporate a budget term $b$ to constrain the network learns to mask out only $b$\% of the total number of pixels in the image. Then, our budget-regularization term $R_b$ given by
\begin{equation*}
    R_b(m, b) = \Big(\sum_{i, j} m_{i, j} - b \Big)^2.
\end{equation*}
We note that this budget constraint is similar to MAE \cite{he2022masked} and other random masking schemes, which choose to randomly mask out a fixed 75\% of the image. However, since our masking procedure is much stronger than random, our selection of $b$ tends to be much smaller. For our experiments, we choose $b = 0.25$, which is roughly equivalent to the strength of masks in ADIOS. To encourage our sequentially generated masks to be disjoint, we introduce an overlap penalty. We can simply compute a dot product between a proposed mask $m^{i}$ with the sum of all previous masks $\sum_{j=1}^{i-1} m^{j}$. This penalty term encourages the newly generated mask to be disjoint from previous masks, and it is a more flexible constraint as compared to that in ADIOS. As a result, our final optimization problem is given by a combination of the encoder loss, the budget regularization, and the overlapping penalty, or
\begin{equation*}
    \mathcal{I}^*, \mathcal{M}^* = \argmin_{\mathcal{I}} \max_{\mathcal{M}} \left( \frac{1}{N} \sum_{i=1}^N \mathcal{L}_{\text{SimCLR}}^{\text{encoder}}(x, \mathcal{I}, m^i) - R_b(m^i, b) - \Big( m^i \cdot \sum_{j=1}^{i-1} m^j \Big) \right).
\end{equation*}
% where $m^i = \mathcal{M}(x, m^1, ..., m^{i-1})$.
% \editcomment{It might be a good idea to be more rigorous when defining the functions and variables: clearly relate L, LSimCLR, and the other regularization terms, to tie it back to the first equation}

\begin{table}[t]
    \centering
    \setlength{\tabcolsep}{10pt}
    {\renewcommand{\arraystretch}{1.3}% for the vertical padding
    \resizebox{0.85\columnwidth}{!}{
    \begin{tabular}{ l | cc  cc }
    \toprule
    & \multicolumn{2}{c}{ImageNet100s} & \multicolumn{2}{c}{STL10} \\ \cmidrule(lr){2-3} \cmidrule(lr){4-5} 
    Method & Linear & Fine Tune
              & Linear & Fine Tune \\
    \midrule
    
    SimCLR & \textit{55.10 $\pm$ 0.15} & - & \textit{85.10 $\pm$ 0.12} & - \\
    + ADIOS & 55.91 $\pm$ 0.12 & 64.83 $\pm$ 0.22 & 86.03 $\pm$ 0.03 & 86.82 $\pm$ 0.11 \\
    + Sequential & \textbf{58.95 $\pm$ 0.10} & \textbf{65.43 $\pm$ 0.34} & \textbf{86.4 $\pm$ 0.04} & \textbf{89.41 $\pm$ 0.26 }  \\
    \bottomrule
    \end{tabular} 
    }
    }
    \vspace{2mm}
    \caption{Results for ID classification (top-1 accuracy) when pretrained and evaluated on ImageNet100s and STL10. Results are averaged over 3 seeds. \textit{Italics} denote results that are reported from existing work \cite{shi2022adversarial}. The best performing method is denoted in bold. }
    \vspace{-3mm}
    \label{tab: ID classification}
\end{table}

\section{Experiments}

We provide empirical results to demonstrate the efficacy of using our sequential mask generation in a SSL pretraining objective. We compare the performance of our method to ADIOS and the base method of SimCLR across classification, transfer learning, and semantic segmentation tasks. For our experiments, we largely follow the same experimental procedure in ADIOS. We evaluate ADIOS and our approach with SimCLR \cite{chen2020simple} as the underlying contrastive SSL objective. For all of these approaches, we use a ResNet18 \cite{He2016DeepRL} backbone. We provide details about our hyperparameters and architectures in Appendix \ref{appx:hyperparam}.

\subsection{Classification Results}
% \subsection{Results \editcomment{when compared to the header of section 3.2 this seems a bit odd}}

We pretrain networks on STL10 and ImageNet100s and then perform classification on the same dataset. STL10 is a dataset consisting of 10 object classes from ImageNet \cite{Russakovsky2015ImageNetLS}, and ImageNet100s is a compressed (96x96 pixels) version of ImageNet100 \cite{Tian2020ContrastiveMC} that is introduced in ADIOS \cite{shi2022adversarial}. 

We report our classification results for both linear probing (LP) and fine tuning (FT) on these datasets in Table \ref{tab: ID classification}. We observe that our sequential mask procedure outperforms ADIOS across all tasks. We highlight that our method significantly improves upon LP of ADIOS on ImageNet100s (by 3 accuracy points or a 7\% reduction in error) and upon FT on STL (by 2.5 accuracy points or almost a 3\% reduction in error). We remark that these are larger improvements than ADIOS observes over the base method SimCLR.   

\begin{table}[t]
    \centering
    \setlength{\tabcolsep}{4pt}
    {\renewcommand{\arraystretch}{1.4}% for the vertical padding
    
    \resizebox{0.99\columnwidth}{!}{
    \begin{tabular}{ l | cc  cc cc | c}
    \toprule
    & \multicolumn{2}{c}{CIFAR10} & \multicolumn{2}{c}{CIFAR100} &  \multicolumn{2}{c}{iNaturalist21} & Pascal VOC\\ \cmidrule(lr){2-3} \cmidrule(lr){4-5} \cmidrule(lr){6-7} \cmidrule(lr){8-8}
    Method & Linear & Fine Tune
              & Linear & Fine Tune & Linear & Fine Tune & Fine Tune \\
    \midrule
    SimCLR & 62.32 $\pm$ 0.21 & 91.34 $\pm$ 0.90 & 37.91 $\pm$ 0.16 & 65.04 $\pm$ 0.48 & 79.78 $\pm$ 0.06 & 92.78 $\pm$ 0.04 & 36.63 $\pm$ 0.42 \\
    + ADIOS & 62.93 $\pm$ 0.42 & 93.93 $\pm$ 0.27 & 38.75 $\pm$ 0.08 & \textbf{73.26 $\pm$ 1.32} & 81.07 $\pm$ 0.02 & 93.74 $\pm$ 0.03 & 38.14 $\pm$ 0.47 \\
    + Sequential & \textbf{63.97 $\pm$ 0.05} & 93.59 $\pm$ 0.33  & 38.83 $\pm$ 0.09 & 72.44 $\pm$ 0.46 & 80.79 $\pm$ 0.02 & 93.31 $\pm$ 0.03 & \textbf{39.86 $\pm$ 0.14} \\
    \bottomrule
    \end{tabular} 
    }}
    \vspace{2mm}
    \caption{Results for classification (top-1 accuracy) on CIFAR10, CIFAR100, and iNaturalist21. Results for semantic segmenatation (mIoU) on Pascal VOC. All methods have been pretrained on ImageNet100s. Results are averaged over 3 seeds, and improvements larger than 1\% are shown in \textbf{bold}.}
    \label{tab: transfer}
    \vspace{-5mm}
\end{table}
% \begin{table}[t]
%     \centering
%     \setlength{\tabcolsep}{10pt}
%     {\renewcommand{\arraystretch}{1.3}% for the vertical padding
%     \resizebox{0.33\columnwidth}{!}{
%     \begin{tabular}{ l | c}
%     \toprule
%     Method & Pascal VOC \\ \midrule
%     SimCLR & 36.63 $\pm$ 0.42 \\
%     + ADIOS & 38.14 $\pm$ 0.47  \\
%     + Sequential & \textbf{39.86 $\pm$ 0.14} \\
%     \bottomrule
%     \end{tabular} 
%     }}
%     \vspace{2mm}
    
%     \caption{Results for semantic segmentation (mIoU) on Pascal VOC, when the network is pretrained on ImageNet100s. Results are averaged over 3 seeds.}
%     \label{tab: segmentation}
% \end{table}

% combine 

\subsection{Transfer Learning and Segmentation Results}

We also provide experimental results on multiple different transfer learning tasks. We first pretrain all methods on ImageNet100s and use these learnt encoders for classification on CIFAR10, CIFAR100, and iNaturalist 2021 \cite{Horn2017TheIC}. We also provide results on a semantic segmentation task on Pascal VOC \cite{everingham2010pascal}. While we only add a linear layer for the classification tasks, we add connections between the ResNet's layers for the segmentation task, following the architectural choices of FCN \cite{Shelhamer2015FullyCN}.

We report the transfer learning results and the semantic segmentation results in Table \ref{tab: transfer}. We observe comparable results across all transfer learning datasets. Our method is slightly better than ADIOS for LP on CIFAR10, but it is slightly weaker than ADIOS for FT on CIFAR100. These results support that our method learns better (more linearly separable) features for these downstream transfer learning tasks. While our method is slightly worse on FT, we note that all these approaches do achieve 100\% training accuracy, so this may be better addressed by modifying the FT procedure with additional regularization. We do remark that our approach is better on the semantic segmentation task by more than 1.5 accuracy points or a 5\% relative improvement, suggesting that our approach learns features that are better suited for more fine-grained reasoning. 

\section{Discussion}

We provide a new framework to sequentially generate masks in an adversarial fashion to enhance the standard contrastive learning pipeline for SSL. Our approach improves upon existing work by changing the underlying mechanism for constraining the adversary and introduces a new penalty to encourage that the learnt masks are disjoint. We demonstrate our method on various experimental tasks, showing significant improvements on ImageNet100s and STL10, as well as in linear readout on downstream classification and in fine-tuning for a semantic segmentation task.  While our method shows improvement for many of these metrics, we expect the most benefit when tackling other, larger datasets. For pretraining on ImageNet100s and STL, we note that most images consist of a single, foreground image, which may not require multiple masks. However, for larger multi-object images (which may potentially not be a majority of the foreground), our approach will likely show even larger benefits over random masking and previous adversarial approaches. Scaling up our adversarial method to larger, multi-object datasets is an open direction for future research.

Finally, we note that our approach is flexible and can benefit from improved approaches for setting the budget constraint on the masking network. We are currently investigating a principled way to determine a dynamic budget via using the sizes of clusters of superpixels (clusters determined by average RGB values) as a proxy for the budget for each mask. This further would strengthen our masking procedure, as it would now be able to mask out exactly the size of particular objects in the image, and this budget would now depend on the input image. We believe that determining this budget in a dynamic fashion is another fertile area of future research.

\bibliographystyle{plain}
\bibliography{main.bib}

\begin{thebibliography}{10}

\bibitem{assran2022masked}
Mahmoud Assran, Mathilde Caron, Ishan Misra, Piotr Bojanowski, Florian Bordes,
  Pascal Vincent, Armand Joulin, Michael Rabbat, and Nicolas Ballas.
\newblock Masked siamese networks for label-efficient learning.
\newblock {\em arXiv preprint arXiv:2204.07141}, 2022.

\bibitem{Bao2022BEiTBP}
Hangbo Bao, Li~Dong, and Furu Wei.
\newblock Beit: Bert pre-training of image transformers.
\newblock {\em ArXiv}, abs/2106.08254, 2022.

\bibitem{Caron2021EmergingPI}
Mathilde Caron, Hugo Touvron, Ishan Misra, Herv'e J'egou, Julien Mairal, Piotr
  Bojanowski, and Armand Joulin.
\newblock Emerging properties in self-supervised vision transformers.
\newblock {\em 2021 IEEE/CVF International Conference on Computer Vision
  (ICCV)}, pages 9630--9640, 2021.

\bibitem{chen2020simple}
Ting Chen, Simon Kornblith, Mohammad Norouzi, and Geoffrey Hinton.
\newblock A simple framework for contrastive learning of visual
  representations.
\newblock In {\em International conference on machine learning}, pages
  1597--1607. PMLR, 2020.

\bibitem{chen2020big}
Ting Chen, Simon Kornblith, Kevin Swersky, Mohammad Norouzi, and Geoffrey~E
  Hinton.
\newblock Big self-supervised models are strong semi-supervised learners.
\newblock {\em Advances in neural information processing systems},
  33:22243--22255, 2020.

\bibitem{chen2020improved}
Xinlei Chen, Haoqi Fan, Ross Girshick, and Kaiming He.
\newblock Improved baselines with momentum contrastive learning.
\newblock {\em arXiv preprint arXiv:2003.04297}, 2020.

\bibitem{everingham2010pascal}
Mark Everingham, Luc Van~Gool, Christopher~KI Williams, John Winn, and Andrew
  Zisserman.
\newblock The pascal visual object classes (voc) challenge.
\newblock {\em International journal of computer vision}, 88(2):303--338, 2010.

\bibitem{gans}
Ian Goodfellow, Jean Pouget-Abadie, Mehdi Mirza, Bing Xu, David Warde-Farley,
  Sherjil Ozair, Aaron Courville, and Yoshua Bengio.
\newblock Generative adversarial networks.
\newblock {\em Communications of the ACM}, 63(11):139--144, 2020.

\bibitem{hamilton2022unsupervised}
Mark Hamilton, Zhoutong Zhang, Bharath Hariharan, Noah Snavely, and William~T
  Freeman.
\newblock Unsupervised semantic segmentation by distilling feature
  correspondences.
\newblock {\em arXiv preprint arXiv:2203.08414}, 2022.

\bibitem{he2022masked}
Kaiming He, Xinlei Chen, Saining Xie, Yanghao Li, Piotr Doll{\'a}r, and Ross
  Girshick.
\newblock Masked autoencoders are scalable vision learners.
\newblock In {\em Proceedings of the IEEE/CVF Conference on Computer Vision and
  Pattern Recognition}, pages 16000--16009, 2022.

\bibitem{he2020momentum}
Kaiming He, Haoqi Fan, Yuxin Wu, Saining Xie, and Ross Girshick.
\newblock Momentum contrast for unsupervised visual representation learning.
\newblock In {\em Proceedings of the IEEE/CVF conference on computer vision and
  pattern recognition}, pages 9729--9738, 2020.

\bibitem{He2016DeepRL}
Kaiming He, X.~Zhang, Shaoqing Ren, and Jian Sun.
\newblock Deep residual learning for image recognition.
\newblock {\em 2016 IEEE Conference on Computer Vision and Pattern Recognition
  (CVPR)}, pages 770--778, 2016.

\bibitem{Horn2017TheIC}
Grant~Van Horn, Oisin~Mac Aodha, Yang Song, Alexander Shepard, Hartwig Adam,
  Pietro Perona, and Serge~J. Belongie.
\newblock The inaturalist challenge 2017 dataset.
\newblock {\em ArXiv}, abs/1707.06642, 2017.

\bibitem{Joshi2020SpanBERTIP}
Mandar Joshi, Danqi Chen, Yinhan Liu, Daniel~S. Weld, Luke Zettlemoyer, and
  Omer Levy.
\newblock Spanbert: Improving pre-training by representing and predicting
  spans.
\newblock {\em Transactions of the Association for Computational Linguistics},
  8:64--77, 2020.

\bibitem{ke2022unsupervised}
Tsung-Wei Ke, Jyh-Jing Hwang, Yunhui Guo, Xudong Wang, and Stella~X Yu.
\newblock Unsupervised hierarchical semantic segmentation with multiview
  cosegmentation and clustering transformers.
\newblock In {\em Proceedings of the IEEE/CVF Conference on Computer Vision and
  Pattern Recognition}, pages 2571--2581, 2022.

\bibitem{kenton2019bert}
Jacob Devlin Ming-Wei~Chang Kenton and Lee~Kristina Toutanova.
\newblock Bert: Pre-training of deep bidirectional transformers for language
  understanding.
\newblock In {\em Proceedings of NAACL-HLT}, pages 4171--4186, 2019.

\bibitem{kim2020unsupervised}
Wonjik Kim, Asako Kanezaki, and Masayuki Tanaka.
\newblock Unsupervised learning of image segmentation based on differentiable
  feature clustering.
\newblock {\em IEEE Transactions on Image Processing}, 29:8055--8068, 2020.

\bibitem{Lan2020ALBERTAL}
Zhenzhong Lan, Mingda Chen, Sebastian Goodman, Kevin Gimpel, Piyush Sharma, and
  Radu Soricut.
\newblock Albert: A lite bert for self-supervised learning of language
  representations.
\newblock {\em ArXiv}, abs/1909.11942, 2020.

\bibitem{Li2022SemMAESM}
Gang Li, Heliang Zheng, Daqing Liu, Bing Su, and Changwen Zheng.
\newblock Semmae: Semantic-guided masking for learning masked autoencoders.
\newblock {\em ArXiv}, abs/2206.10207, 2022.

\bibitem{Liu_2017_ICCV}
Shu Liu, Jiaya Jia, Sanja Fidler, and Raquel Urtasun.
\newblock Sgn: Sequential grouping networks for instance segmentation.
\newblock In {\em Proceedings of the IEEE International Conference on Computer
  Vision (ICCV)}, Oct 2017.

\bibitem{pathak2016context}
Deepak Pathak, Philipp Krahenbuhl, Jeff Donahue, Trevor Darrell, and Alexei~A
  Efros.
\newblock Context encoders: Feature learning by inpainting.
\newblock In {\em Proceedings of the IEEE conference on computer vision and
  pattern recognition}, pages 2536--2544, 2016.

\bibitem{Ronneberger2015UNetCN}
Olaf Ronneberger, Philipp Fischer, and Thomas Brox.
\newblock U-net: Convolutional networks for biomedical image segmentation.
\newblock {\em ArXiv}, abs/1505.04597, 2015.

\bibitem{Russakovsky2015ImageNetLS}
Olga Russakovsky, Jia Deng, Hao Su, Jonathan Krause, Sanjeev Satheesh, Sean Ma,
  Zhiheng Huang, Andrej Karpathy, Aditya Khosla, Michael~S. Bernstein,
  Alexander~C. Berg, and Li~Fei-Fei.
\newblock Imagenet large scale visual recognition challenge.
\newblock {\em International Journal of Computer Vision}, 115:211--252, 2015.

\bibitem{Shelhamer2015FullyCN}
Evan Shelhamer, Jonathan Long, and Trevor Darrell.
\newblock Fully convolutional networks for semantic segmentation.
\newblock {\em 2015 IEEE Conference on Computer Vision and Pattern Recognition
  (CVPR)}, pages 3431--3440, 2015.

\bibitem{shi2022adversarial}
Yuge Shi, N~Siddharth, Philip Torr, and Adam~R Kosiorek.
\newblock Adversarial masking for self-supervised learning.
\newblock In {\em International Conference on Machine Learning}, pages
  20026--20040. PMLR, 2022.

\bibitem{Tian2020ContrastiveMC}
Yonglong Tian, Dilip Krishnan, and Phillip Isola.
\newblock Contrastive multiview coding.
\newblock In {\em ECCV}, 2020.

\bibitem{xu2022groupvit}
Jiarui Xu, Shalini De~Mello, Sifei Liu, Wonmin Byeon, Thomas Breuel, Jan Kautz,
  and Xiaolong Wang.
\newblock Groupvit: Semantic segmentation emerges from text supervision.
\newblock In {\em Proceedings of the IEEE/CVF Conference on Computer Vision and
  Pattern Recognition}, pages 18134--18144, 2022.

\end{thebibliography}

\clearpage

\appendix

\section*{Appendix}

% Optionally include extra information (complete proofs, additional experiments and plots) in the appendix.
% This section will often be part of the supplemental material.

\section{Related Work}

\textbf{Contrastive Learning}

Many SSL pretext tasks fall under the category of contrastive learning, where a network learns an invariance to a particular set of data augmentations. This line of work \citep{chen2020simple, he2020momentum, chen2020big, chen2020improved, assran2022masked} trains a network by making two positive views (i.e., differently augmented views of the \textit{same} image) nearby in a latent space, while making two negative views (i.e., augmented version of \textit{different} images) far apart. However, this approach can be computationally expensive as it requires a large batch size to have sufficient negative samples. Also, relying on strong image augmentations can potentially be harmful for particular downstream tasks; for example, relying on rotations and flips might harm the network's ability to classify directions.

\textbf{Masked-based SSL}

More recently, other approaches have applied mask-based reconstruction tasks in SSL. These approaches \citep{he2022masked,assran2022masked, Bao2022BEiTBP} have demonstrated strong performance albeit using a random masking procedure. Under a random masking scheme, a large amount of pixels (50-75\% depending on the architecture) need to be masked to encourage a network to learn some notion of global reasoning, which benefits downstream tasks. Other works have shown strong initial progress in improving this masking procedure to learn more meaningful, difficult masks \citep{shi2022adversarial, Li2022SemMAESM}. ADIOS \citep{shi2022adversarial} is the most related work, which looks to learn a masking model in an adversarial fashion with particular constraints on the adversary. This work also empirically demonstrates that better masks (i.e., masks that correspond to individual object boundaries) lead to better (ID) classification performance. Our work looks to further this line of work by changing the procedure that existing work uses to generate masks and changing the underlying adversary constraints. 

\textbf{Unsupervised Segmentation}

While not directly related to learning representations for classification, there are similarities between learning how to mask out objects and unsupervised segmentation tasks. The goal of unsupervised segmentation is to group regions of images together that capture objects or other semantic meaning without any (pixel-level) supervision. Many existing works \citep{kim2020unsupervised, Caron2021EmergingPI, xu2022groupvit, hamilton2022unsupervised, ke2022unsupervised} tackle this problem by encouraging consistency or grouping across input tokens to vision transformers. An adversarially learnt masking network seems to learn some unsupervised segmentation capability, although it is trained for a fundamentally different objective without any of these loss terms to encourage consistency. As a result, adversarially learnt masks are semantically meaningful but are imperfect as they sometimes do not respect object boundaries or contain various noisy pixels. 
% \editcomment{Not sure if it's true that the adversarially learned masks don't respect object boundaries at all - it's rough, but looking by definition the task of segmentation is to respect object boundaries while dividing the image. if this is not the argument, please consider making this clearer}

% cite various grouping works
\section{Mask Visualizations} \label{app:vis_mask}

\begin{figure}[t]
    \centering
    \begin{subfigure}[t]{0.49\textwidth}
        \centering
        \includegraphics[width=\textwidth]{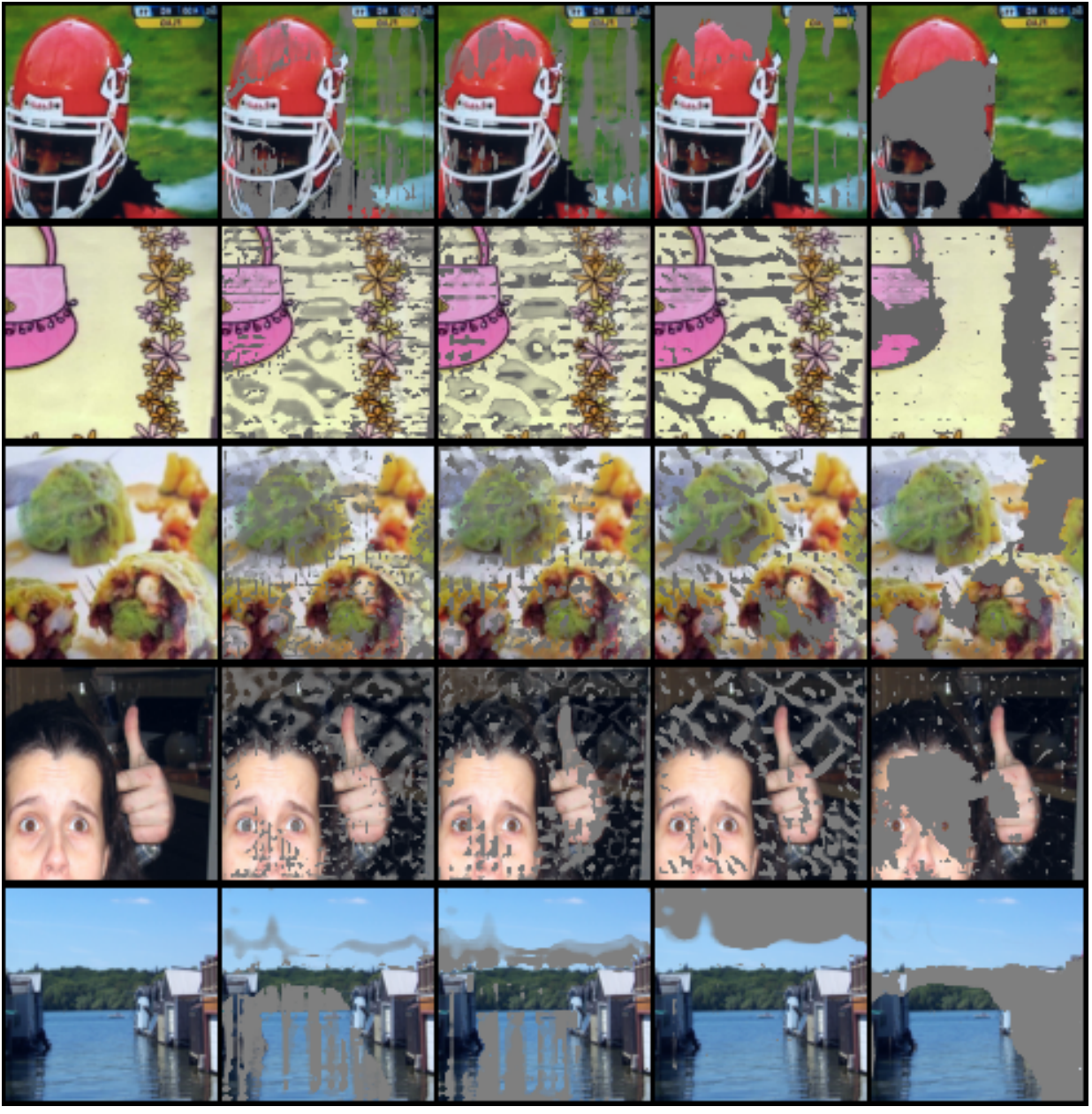}
        \caption{Masks from ADIOS}
    \end{subfigure}
    \hfill
    \begin{subfigure}[t]{0.49\textwidth}
        \centering
        \includegraphics[width=\textwidth]{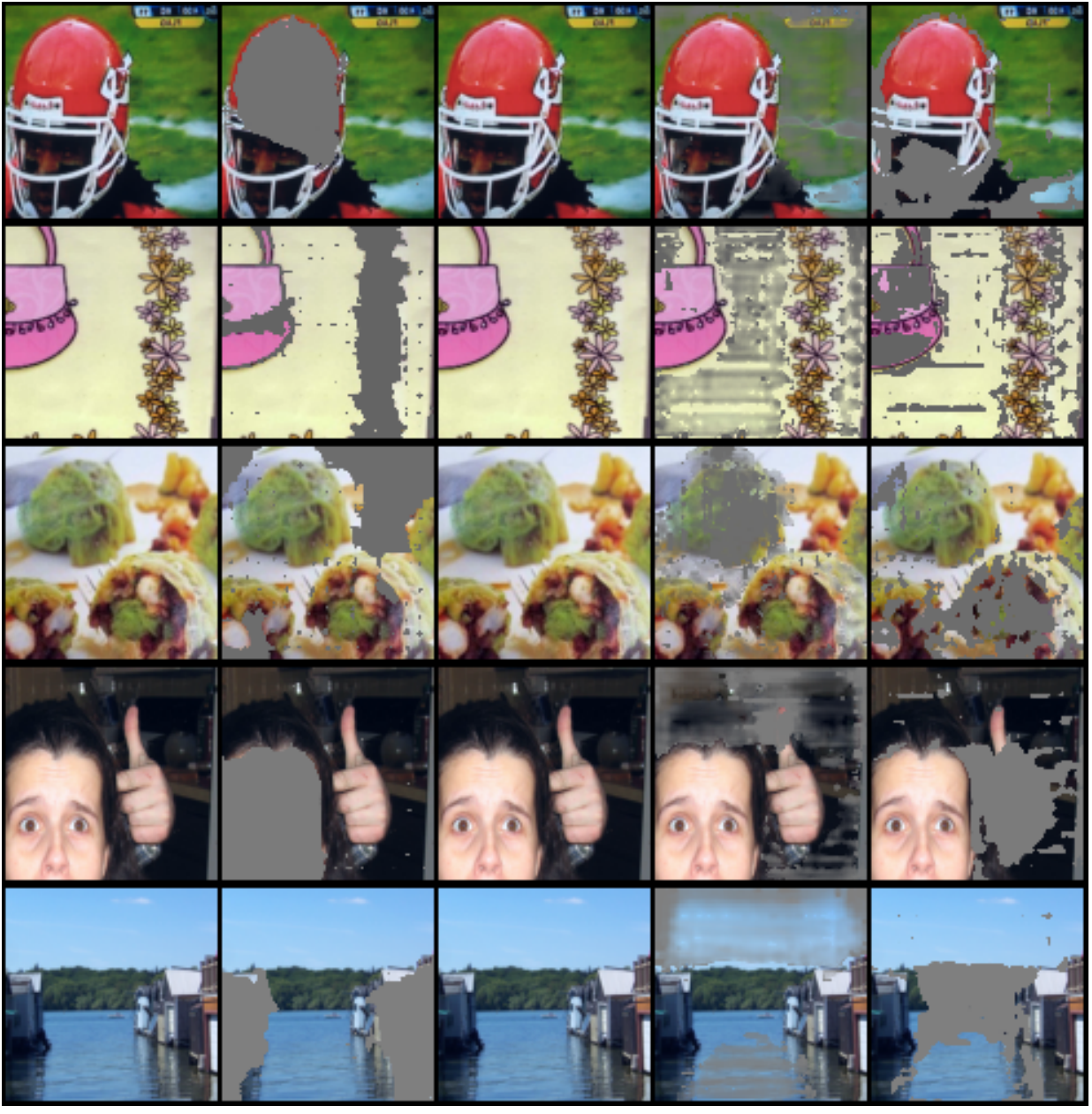}
        \caption{Masks from our sequential masking framework}
    \end{subfigure}
    \caption{Visualization of the masks generated by ADIOS (a) and our sequential mask pipeline (b) on ImageNet100s images. For ADIOS, we use $N = 4$, and for our pipeline, we use $N = 4$ and $b=0.25$. Within each row, the leftmost image is the original image, and the remaining images correspond to the $N = 4$ learnt masks. We note that our method learns to mask out nothing for one output as the penalty incurred by overlapping and consistency outweighs the increased loss from a new masked region.}
    \label{fig:mask_vis}
\end{figure}

We provide visualizations of the masks generated by ADIOS in Figure \ref{fig:mask_vis} (a) and by our sequential mask generation procedure in Figure \ref{fig:mask_vis} (b). We observe that both masking procedures frequently learn to remove semantically meaningful regions to pass into the encoder network. For ADIOS, in many instances, the rightmost masking channel learns to remove one continuous foreground object, while the remaining masks learn to fight over noise and patterns in the background. In our sequential masking procedure, we observe that our network learns to produce multiple semantically meaningful masks. On ImageNet100s, we observe that 3 of the 4 masks learn meaningful regions in the image. The leftmost mask corresponds to the foreground object. Two other masks learn to mask out parts of the foreground and the background. Finally, the last mask learns to mask out nothing; the penalty incurred by removing any more regions (and overlapping with existing masks) is larger than the increased loss of the encoder network. We can observe that in a few instances (e.g., the second row and the fourth row) our sequential masking network seems to learn masks that better correspond to the multiple foreground objects in the image, such as the person's face and hand. On the other hand, multiple masks from ADIOS learn to remove noise or patterns in the background. This finding supports the added flexibility of our method; with ADIOS, each network must remove roughly equal amounts of pixels and each pixel in the image must completely be removed. However, with our more flexible constraint, we can develop better masking procedures, as is demonstrated on this dataset by only requiring 3 meaningful masks for better performance. 
% \editcomment{if additional visualizations from ADIOS can be added here (with officially released ADIOS model) we should be able to also show that ADIOS has a difficult time sorting different regions of the image consistently into individual channels - recall the issue where ADIOS' masking network seems to only have a single active channel as a byproduct of their CNN architecture}

We note that including this fourth mask (and other additional masks) seems to further increase performance; it serves a purpose similar to regularization towards the original contrastive method (SimCLR), by adding additional positive augmented pairs without any masking. We also note that adding a small consistency penalty (Appendix \ref{consistency}) encourages our network to learn more continuous, well-grouped regions. This further discourages that each mask needs to remove a region; the remaining regions after some masks may not be continuous, and thus a masking network incurs less penalty by not producing any mask. In all, both masking procedures are still imperfect, although our sequential framework seems to be a step towards generating more meaningful segmentations of the original image.

\section{Consistency Penalty} \label{consistency}

We observe that the masks that are learnt by existing work \citep{shi2022adversarial} are imperfect and contain noise in many masks. We can improve the quality of these masks by introducing a small consistency penalty, which encourages the network to remove more continuous regions. Our penalty takes the form of 
\begin{equation*}
    C(m) = || m - m^* ||_2^2
\end{equation*}
where $m$ is a predicted mask, and $m^*$ is an average-pooled version of the same mask. By encouraging a mask to be similar to its average-pooled version, we are encouraging the masks to be smooth and not have different pixel values within a small neighborhood defined by our average-pooling kernel. For our experiments, we use a kernel size of 3x3. 

\begin{table}[t]
    \centering
    \setlength{\tabcolsep}{4pt}
    {\renewcommand{\arraystretch}{1.4}% for the vertical padding
    \resizebox{0.4\columnwidth}{!}{
    \begin{tabular}{cc}
    \toprule 
        Name & Value \\ \midrule
        Optimizer & SGD \\
        Momentum & 0.9 \\
        Scheduler & warmup cosine \\
        Epochs & 500 \\
        Batch Size & 256 \\ 
        Enc. Learning Rate & 0.11 \\
        Temperature & 0.2 \\
        \bottomrule
    \end{tabular} 
    }}
    \vspace{2mm}
    \caption{Pretraining hyperparameters used for all methods}
    \label{hyp_all}
    % \vspace{-5mm}
\end{table}

\section{Parameter and Architecture Settings} \label{appx:hyperparam}

For all of the methods, we use a ResNet18 \citep{He2016DeepRL} as an encoder and a single feed-forward layer as our projection head for SimCLR. For the masking network, we use a U-Net \citep{Ronneberger2015UNetCN} for the mask generation network, which is shared among all mask heads. We use 1x1 convolutional layers to generate our $N$ masks. For our optimization and training, we follow many of the same parameter settings that are used in ADIOS \citep{shi2022adversarial}.

\begin{table}[t]
    \centering
    \setlength{\tabcolsep}{4pt}
    {\renewcommand{\arraystretch}{1.4}% for the vertical padding
    \resizebox{0.42\columnwidth}{!}{
    \begin{tabular}{cc | cc}
    \toprule 
    Name & Value & Name & Value \\ \midrule
    N & 4 & N & 5\\ 
    entropy & 0.71 & budget & 0.25 \\ 
    sparsity & 0.93 & overlap & 0.0001 \\
    & & consistency & 0.0001 \\   \bottomrule
    \end{tabular} 
    }
    }
    \vspace{2mm}
    \caption{Pretraining hyperparameters used for ADIOS (left) and our sequential masking procedure (right).}
    \label{hyp_adios_ours}
\end{table}

We note that we use $N = 5$ and $b = 0.25$ in our pipeline. This results in two different masks learning to mask out nothing (i.e., adding two additional regularizing pairs of positive images from standard SimCLR). As noted above, this leads to slightly better performance, further improving upon the gains from our sequential masking procedure. 

\section{Downstream Task Setup}

For our linear probe and fine tuning experiments on classification tasks, we run all approaches with a fixed learning rate of 0.1, a batch size of 256, and no weight decay. For fine tuning, we use a learning rate of 0.5, a batch size of 256, and no weight decay. 

% \begin{table}[h]
%     \centering
%     \setlength{\tabcolsep}{4pt}
%     {\renewcommand{\arraystretch}{1.4}% for the vertical padding
%     \resizebox{0.22\columnwidth}{!}{
%     \begin{tabular}{cc}
%     \toprule 
%         Name & Value \\ \midrule
%         N & 5 \\ 
%         budget & 0.25 \\
%         consistency & 0.0001\\
%         overlap &  0.0001 \\ \bottomrule 
%     \end{tabular} 
%     }
%     }
%     \vspace{2mm}
%     \caption{Pretraining hyperparameters used for our sequential masking procedure.}
%     \label{tab: transfer}
%     \vspace{-5mm}
% \end{table}

% \section{Importance of Additional Contrastive Pairs}

\end{document}